\documentclass{article} 

\usepackage[preprint]{colm2026_conference}

\usepackage{microtype}
\usepackage{hyperref}
\usepackage{url}
\usepackage{booktabs}

\usepackage{lineno}

\definecolor{darkblue}{rgb}{0, 0, 0.5}
\hypersetup{colorlinks=true, citecolor=darkblue, linkcolor=darkblue, urlcolor=darkblue}

\usepackage{tikz}
\usepackage{pgfplots}
\usepgfplotslibrary{colormaps}
\usepgfplotslibrary{colorbrewer}
\usepackage{siunitx}
\usepackage{subcaption}
\usepackage{amsmath}
\usepackage{graphicx}
\usepackage{tabularx}
\usepackage{ragged2e}

\pgfplotsset{compat=1.18} 

\pgfplotsset{
  seaborn/.style={
    axis lines=left,
    axis line style={black,-},
    tick style={black},
    tick align=outside,
    tick label style={font=\small},
    label style={font=\small},
    title style={font=\small},
    legend style={draw=none, fill=none, font=\small},
    grid=both,
    major grid style={line width=0.3pt, draw=black!12},
    minor grid style={line width=0.2pt, draw=black!7},
    minor tick num=1,
    axis background/.style={fill=white},
  },
  seaborn lines/.style={
    every axis plot/.append style={thin, mark=*, mark size=2pt, opacity=0.8}
  }
}
\pgfplotsset{
    colormap={viridis_r}{
        indices of colormap={
            \pgfplotscolormaplastindexof{viridis},...,0 of viridis
        }
    }
}

\title{MoE Routing Testbed: Studying Expert Specialization and Routing Behavior at Small Scale}

\author{\textbf{Tobias Falke$^{*}$}{\hspace{.1em}}
    \quad
    \textbf{Nicolas Anastassacos$^{*}$}{\hspace{.1em}}
    \quad
    \textbf{Samson Tan}{\hspace{.1em}}
    \quad
    \textbf{Chankrisna Richy Meas}{\hspace{.1em}}
    \quad
    \vspace{.2em}\\
    \textbf{Chandana Satya Prakash}{\hspace{.1em}}
    \quad
    \textbf{Nitesh Sekhar}{\hspace{.1em}}
    \quad
    \textbf{M Saiful Bari}{\hspace{.1em}}
    \quad
    \textbf{Krishna Kompella}{\hspace{.1em}}
    \quad
    \vspace{.2em}\\
    \textbf{Gamaleldin F. Elsayed$^{*}$}{\hspace{.1em}}
    \quad
    \vspace{.5em}\\
    Amazon AGI
    \vspace{.5em}\\
    \texttt{falket@amazon.com}
}

\begin{document}

\definecolor{red1}{RGB}{34,19,48}
\definecolor{red2}{RGB}{68,27,70}
\definecolor{red3}{RGB}{105,31,85}
\definecolor{red4}{RGB}{145,28,91}
\definecolor{red5}{RGB}{184,22,86}
\definecolor{red6}{RGB}{217,39,71}
\definecolor{red7}{RGB}{237,80,62}
\definecolor{red8}{RGB}{243,125,86}
\definecolor{red9}{RGB}{245,164,123}
\definecolor{red10}{RGB}{246,200,170}

\pgfplotscreateplotcyclelist{custom}{
  {Paired-A, solid, mark=*},
  {Paired-B, dashed, mark=square*},
  {Paired-C, dotted, mark=triangle*},
  {Paired-D, dashdotted, mark=diamond*},
  {Paired-E, densely dashed, mark=pentagon*},
  {Paired-F, densely dotted, mark=otimes*},
  {Paired-G, loosely dashed=star},
  {Paired-H, loosely dotted=oplus*},
}

\pgfplotscreateplotcyclelist{custom2}{
  {blue, solid, mark=*},
  {cyan, dashed, mark=*},
  {orange, dotted, mark=*},
  {red5, dashdotted, mark=*},
  {blue, dashed, mark=*},
  {cyan, dotted, mark=*},
  {orange, dashdotted, mark=*},
  {red5, dashed, mark=*},
}

\definecolor{tab0}{HTML}{1F77B4}
\definecolor{tab1}{HTML}{FF7F0E}
\definecolor{tab2}{HTML}{2CA02C}
\definecolor{tab3}{HTML}{D62728}
\definecolor{tab4}{HTML}{9467BD}
\definecolor{tab5}{HTML}{8C564B}
\definecolor{tab6}{HTML}{E377C2}
\definecolor{tab7}{HTML}{7F7F7F}
\definecolor{tab8}{HTML}{BCBD22}
\definecolor{tab9}{HTML}{17BECF}

\pgfplotscreateplotcyclelist{tab10}{
  {black, solid},
  {tab0, dashed},
  {tab1, dashed},
  {tab2, dashed},
  {tab3, dashed},
  {tab4, dashed},
  {tab5, dashed},
  {tab6, dashed},
  {tab7, dashed},
  {tab8, dashed},
  {tab9, dashed},
}

\ifcolmsubmission
\linenumbers
\fi

\maketitle

\renewcommand{\thefootnote}{\fnsymbol{footnote}}
\footnotetext[1]{Core contributors}
\renewcommand{\thefootnote}{\arabic{footnote}}

\begin{abstract}
Sparse Mixture-of-Experts (MoE) architectures are increasingly popular for frontier large language models (LLM) but they introduce training challenges due to routing complexity. Fully leveraging parameters of an MoE model requires all experts to be well-trained and to specialize in non-redundant ways. Assessing this, however, is complicated due to lack of established metrics and, importantly, many routing techniques exhibit similar performance at smaller sizes, which is often not reflective of their behavior at large scale. To address this challenge, we propose the MoE Routing Testbed, a setup that gives clearer visibility into routing dynamics at small scale while using realistic data. The testbed pairs a data mix with clearly distinguishable domains with a reference router that prescribes ideal routing based on these domains, providing a well-defined upper bound for comparison. This enables quantifiable measurement of expert specialization. To demonstrate the value of the testbed, we compare various MoE routing approaches and show that balancing scope is the crucial factor that allows specialization while maintaining high expert utilization. We confirm that this observation generalizes to models 35x larger.
\end{abstract}

\section{Introduction}
\label{sec:introduction}

In recent years, sparse Mixture-of-Experts (MoE) architectures \citep{shazeer2017} have become increasingly popular for frontier large language models (LLMs) such as Gemini \citep{comanici2025gemini25pushingfrontier}, DeepSeekV3 \citep{deepseekai2025deepseekv3technicalreport} and Qwen3 \citep{yang2025qwen3technicalreport}. Their sparse nature enables scaling model capacity without proportional increases in computation, improving training and inference efficiency and allowing model size to grow to scales otherwise impossible \citep{fedus2022switch,deepseekai2025deepseekv3technicalreport}. However, compared to dense architectures, MoEs are harder to train as their routing mechanisms introduce additional hyper-parameters that require careful tuning. Furthermore, perplexity-based metrics may not reveal meaningful differences between routing techniques when token budget and model size are small, posing a risk when scaling up.

Expert specialization has been identified as essential for leveraging the increased capacity of sparse MoE models and preventing learning redundant knowledge representations \citep{dai-etal-2024-deepseekmoe,guo2025advancingexpertspecializationbetter,lv2025couplingexpertsroutersmixtureofexperts}. Yet, despite various attempts to measure expert specialization, so far no widely accepted metric has been established.

\begin{figure}[t]
\begin{minipage}[t]{0.475\textwidth}
    \vskip 0pt
    \centering
    \includegraphics[width=0.825\textwidth]{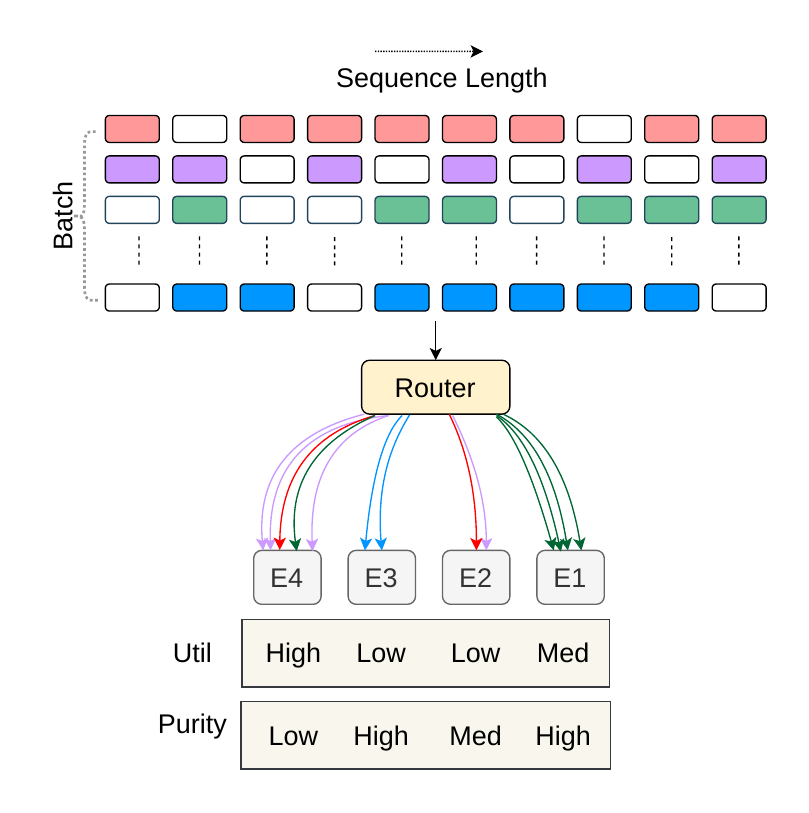}
    \caption{Testbed design: Domains (color-coded) in the data mix are used to define a reference routing for domain-specific tokens (but not generic tokens). For learned routers, the domain purity of tokens sent to a specific expert can then capture the degree of specialization, along with its utilization.} 
    \label{fig:testbed}
\end{minipage}\hfill
\begin{minipage}[t]{0.475\textwidth}
    \vskip -1pt
    \centering

\begin{tikzpicture}
\begin{axis}[
    seaborn, seaborn lines,
    xlabel={Expert Utilization},
    ylabel={Expert Specialization},
    ylabel style={yshift=-3pt},
    grid=major,
    width=1\textwidth,
    xmin=0.5, xmax=1.0,
    ymin=0.15, ymax=0.8,
    legend style={font=\tiny,draw=none,fill=none},
    legend cell align=left,
    legend pos=south west,
]
\addlegendimage{empty legend}
\addlegendentry{\textbf{Scope}}

\addplot[
    color=red1,
    mark=*, mark size=1pt, line width=1pt] table [x=utilization,  y=specialization, col sep=tab] {
utilization	specialization
0.9194703875926504	0.21702174901178
0.9236557245254516	0.2173533856868743
0.9260132789611816	0.2194726079702377
0.924898076057434	0.2236411333084106
0.9199808100962268	0.2317338907811307
0.9101481914520264	0.2448902301490306
0.8811101913452148	0.2855634644627571
0.8163403272628784	0.3260409906506538
0.7510543305386779	0.4162098417679469
0.6825475096702576	0.5096373058855533
0.6405556201934814	0.5703440457582474
0.6049635767936706	0.625653301179409
0.6552371740341186	0.6518066108226777
0.6534011895174807	0.6838555354528476
0.6242419481277466	0.7018988579511642
};
\addlegendentry{1 sequence}
\addplot [only marks, mark=triangle*, mark size=4pt, mark options={fill=red1,draw opacity=0}, forget plot]
  coordinates {(0.9195, 0.2170)};

\addplot[
    color=red3,
    mark=*, mark size=1pt, line width=1pt] table [x=utilization,  y=specialization,  col sep=tab] {
utilization	specialization
0.9186094999313354	0.2126320630311965
0.9300195455551148	0.2207448035478591
0.924139165878296	0.2407414585351943
0.9157280683517456	0.2719502657651901
0.910643458366394	0.3023139059543609
0.9145861148834228	0.3399298444390297
0.8991757869720459	0.3827997699379921
0.8404083251953125	0.4183408617973327
0.7823562860488892	0.4645071968436241
0.7125742197036743	0.5354634463787079
0.64833984375	0.604585325717926
0.5945376873016357	0.66240174472332
0.6456032752990722	0.6845864593982697
0.5688811779022217	0.7189984589815139
0.5768323183059693	0.7389093607664108
};
\addlegendentry{2 sequences}
\addplot [only marks, mark=triangle*, mark size=4pt, mark options={fill=red3,draw opacity=0}, forget plot]
  coordinates {(0.9186, 0.2126)};

\addplot[
    color=red5,
    mark=*, mark size=1pt, line width=1pt] table [x=utilization,  y=specialization,  col sep=tab] {
utilization	specialization
0.914502489566803	0.4199620425701141
0.9064197182655336	0.4879632204771041
0.9113308191299438	0.5214754521846772
0.9063648496355328	0.5604981541633606
0.9057435677878222	0.5708103919748085
0.8997695207595825	0.6274220660328865
0.8760005950927734	0.654454481601715
0.8566638112068177	0.6376988023519516
0.7981155633926391	0.6646485030651093
0.71527179479599	0.6497414857149124
0.6427917957305909	0.6640345811843872
0.6605739116668701	0.6811786741018295
0.6802178144454956	0.6857365399599076
0.6868236780166626	0.6951585695147514
0.5957499504089355	0.714538961648941
};
\addlegendentry{8 sequences}
\addplot [only marks, mark=triangle*, mark size=4pt, mark options={fill=red5,draw opacity=0}, forget plot]
  coordinates {(0.9145, 0.4200)};

\addplot[
    color=red7,
    mark=*, mark size=1pt, line width=1pt] table [x=utilization,  y=specialization,  col sep=tab] {
utilization	specialization
0.9310190200805664	0.4580606520175934
0.9190630435943604	0.5064580574631691
0.9190193176269532	0.5339528173208237
0.9132253646850584	0.5529825061559677
0.9092774152755736	0.6011289536952973
0.9058802843093872	0.6285961389541626
0.8866788864135742	0.6129980325698853
0.8595752000808716	0.621398076415062
0.8272663593292237	0.6220572233200073
0.755805778503418	0.6361255824565888
0.6498228549957276	0.6646367400884629
0.6328745603561401	0.677056896686554
0.613631296157837	0.6999910026788712
0.6148867130279541	0.7207782417535782
0.6213740110397339	0.7315173774957657
};
\addlegendentry{16 sequences}
\addplot [only marks, mark=triangle*, mark size=4pt, mark options={fill=red7,draw opacity=0}, forget plot]
  coordinates {(0.9310, 0.4581)};

\addplot [only marks, mark=pentagon*, mark size=4pt, mark options={fill=black,draw opacity=0}, forget plot]
  coordinates {(0.5727, 0.7295)};

\end{axis}
\end{tikzpicture}
    \caption{The routing testbed can clearly capture the utilization vs. specialization trade-off inherent to MoE training even at small scale. As fewer balancing is applied (\protect\tikz[scale=1.7]\protect\pgfuseplotmark{triangle*}; to \protect\tikz[scale=1.75]\protect\pgfuseplotmark{pentagon*};), utilization is traded off for specialization. Better trade-offs are possible when balancing over more token sequences.}
    \label{fig:teaser}
\end{minipage}
\end{figure}

To address these challenges, we introduce the \textbf{MoE Routing Testbed}, which consists of two key components: a highly separated data mix composed of distinct domains and a reference router created for that mix (Figure~\ref{fig:testbed}). We verify that this reference router outperforms all learned routing methods tested, providing an upper bound for quantitative comparison. We can identify experts that receive all their tokens from a single domain as highly specialized, whereas experts receiving cross-domain tokens as less specialized. We refer to this property as \textit{routing purity}. The testbed provides three main advantages: (1) it enables directly measuring expert specialization through routing purity, (2) it uses realistic natural language data rather than synthetic sequences, staying close to the real-world language modeling task, and (3) can be used reliably at small scale.

We use the routing testbed to study load balancing methods in MoE training. Balancing mechanisms serve two purposes: enabling efficient expert parallelism and preventing router collapse. Yet both can conflict with specialization. Various balancing methods have been proposed, including auxiliary losses \citep{lepikhin2021gshard}, expert biases \citep{wang2024expertbiases}, and assignment algorithms \citep{lewis2021base,clark2022}. As Figure~\ref{fig:teaser} shows, our testbed clearly captures that decreasing balancing strength trades expert utilization for specialization. A crucial factor is the scope of tokens (or sequences) over which balance is encouraged \citep{qiu-etal-2025-demons}. In our testbed, this effect is clearly visible: as scope increases, we are able to achieve better utilization-specialization trade-offs. All data points in the figure were obtained with small-scale models of only 50M active (270M total) parameters.
Beyond the crucial role of scope, we find that allowing sufficient tokens-per-expert at local scope (and thereby avoiding token dropping) is necessary to benefit from specialization. The choice of balancing method itself in fact plays a minor role, all methods perform well at sufficiently large scope. 
Furthermore, we verify that findings from our efficient small scale setup also hold at larger scale (270M $\rightarrow$ 9.6B, 35x larger).

To summarize, we make the following contributions: 1) We introduce the MoE Routing Testbed, enabling discovery of effective routing configurations from small-scale experiments. 2) The testbed directly measures expert specialization and utilization, providing insights beyond standard perplexity metrics. 3) We show that balancing scope matters more than method choice, and that sufficient local tokens-per-expert is critical for specialization. 4) We demonstrate that routing decisions identified at small scale generalize to larger models.

\section{Background}
\label{sec:background}

\subsection{Mixture-of-Experts}

Transformer-based MoE models replace each feed-forward MLP block with $E$ MLP experts, activating only a subset of experts for each token. Given a token representation $\mathbf{x} \in \mathbb{R}^d$ (from a batch of $T$ tokens), a routing network $R: \mathbb{R}^d \to \mathbb{R}^E$ computes routing logits $\mathbf{z} = R(\mathbf{x})$. In top-$k$ token choice routing, the $k$ highest-scoring experts are selected per token and their corresponding logits are normalized using a softmax to obtain routing weights, $w(x)$. The final output is computed as a weighted combination of the selected expert outputs, $\mathbf{y} = \sum_{i \in \text{top-k}(\mathbf{x})} w_i(\mathbf{x}) \cdot E_i(\mathbf{x})$.

\subsection{Load Balancing}
\label{sec:loadbalancing}

Training MoE models typically requires mechanisms to balance token assignments across experts. Without such mechanisms, routers tend to collapse, directing most tokens to a small subset of experts, which can significantly degrade both model quality and efficiency.

\subsubsection{Load Balancing Methods}
We compare four token balancing approaches from the literature: \textbf{Load Balancing Loss (LBL)} \citep{lepikhin2021gshard,fedus2022switch} adds an auxiliary loss $L_{\text{LBL}} = E \cdot \sum_i^E f_i \cdot P_i$ to penalize uneven distributions; \textbf{Balanced Assignment (BA)} \citep{lewis2021base} uses an auction algorithm to enforce exactly $T/E$ tokens per expert; \textbf{Sinkhorn Routing (SH)} \citep{clark2022} applies optimal transport with iterative normalization for approximate balance; and \textbf{Expert Bias (EB)} \citep{deepseekai2025deepseekv3technicalreport} adjusts expert-specific biases based on utilization to converge towards balance over time. These methods differ in whether they enforce strict vs. approximate balance, affect routing immediately vs. with delay, and offer explicit strength control. See Appendix \ref{app:balancing_methods} for detailed descriptions.

\subsubsection{Balancing Scope}
\label{sec:balancing_scope}
In addition to the balancing method and the strength with which it is applied, a third important factor is the scope of tokens $T$ over which balancing is enforced or encouraged. Tokens can be selected across sequence and batch dimensions. If the scope is small, such as spanning just a single token sequence from one data source, applying balancing can introduce a direct conflict with expert specialization since closely related tokens are forced to be routed to different experts. Our experiments in Section~\ref{sec:experiments} clearly demonstrate this.

As recently pointed out by \citet{qiu-etal-2025-demons}, many publicly available training frameworks, by default, balance over the tokens locally present on one device. In large-scale distributed training, where gradient accumulation or pipeline parallelism are commonly used, the per-device micro-batch often contains just a single sequence, making expert specialization challenging. It is therefore important to carefully control the scope of token balancing, especially when also applying sequence parallelism, and to add cross-device communication if needed.
Controlling scope is also crucial for fair comparisons of balancing methods. EB was originally claimed to be superior to LBL \citep{deepseekai2025deepseekv3technicalreport}, but later shown to be on par when used at the same scope \citep{qiu-etal-2025-demons}. Our experiments in the routing testbed explicitly control scope and confirm this finding using only a fraction of the compute.

The conflict between specialization and single-sequence balancing assumes a (packed) sequence is from a single source. Mixing multiple sources into a single training sequence can reduce this challenge, but only to a limited extent. See \ref{app:packing} for a more detailed discussion.

\subsection{Expert Specialization}
\label{sec:specialization}

While early work showed that experts naturally specialize to well-defined clusters in synthetic data \citep{chen2022}, measuring specialization in MoEs trained on real-world data remains challenging. Early approaches compared token distributions across domains \citep{jiang2024mixtralexperts} or measured the impact of excluding top experts \citep{dai-etal-2024-deepseekmoe}. Recent work has developed more precise metrics: routing consistency and score variance \citep{guo2025advancingexpertspecializationbetter}, expert activation norm matrices \citep{lv2025couplingexpertsroutersmixtureofexperts} and expert output similarities \citep{omi2025loadbalancingmixtureexperts}, yet these works struggle with interpretability of their metrics. Most relevant to our work, \citet{dikkala-etal-2023-benefits} use synthetic data with well-defined specialization as a reference for comparing routing approaches. However, whether findings from synthetic data transfer to real language distributions remains unclear.

\section{Routing Testbed}
\label{sec:testbed}

A core challenge in evaluating MoE training is the absence of ground truth for optimal routing decisions. We address this by introducing a routing testbed with two key components: (a) a data mix $\mathcal{M}$ composed of $D$ clearly separable domains $\mathcal{M} = \mathcal{M}_1 \cup \mathcal{M}_2 \cup \dots \cup \mathcal{M}_D$, and (b) a \textbf{reference router} that prescribes ideal expert assignments based on these domains. We verify that this reference router outperforms all learned routing methods tested and, in that sense, provides an upper bound for quantitative comparison.

The reference router enables direct measurement of two critical properties. First, \textbf{expert specialization} can be quantified by comparing learned routing assignments against the reference: experts receiving tokens primarily from a single domain are highly specialized, while those receiving cross-domain tokens are less specialized. Second, \textbf{expert utilization} measures whether all experts receive a comparable number of tokens. While domains provide a strong prior for specialization, not all tokens are equally domain-specific—some appear across all domains. The reference router accounts for this by treating tokens differently based on their domain-specificity (see Section~\ref{sec:split}). The following sections discuss these components of the testbed in detail.

\subsection{Data Mix Construction}

\subsubsection{Desiderata}
A data mix for the routing testbed needs to satisfy the following requirements:

\begin{itemize}
\item \textbf{R1 - Separate Domains:} The mix needs to have clearly defined domains against which routing can be evaluated. This separation needs to exist even at the token level, since routing in MoEs happens per token.
\item \textbf{R2 - Natural Text:} To make the testbed as close as possible to the target task, which is general language modeling on large-scale data, we require the data to be natural human language data, rather than artificial or synthetic data.
\item \textbf{R3 - Sufficient Volume:} We need a sufficient amount of data to train language models on the mix to obtain interpretable results.
\end{itemize}

R3 is easy to satisfy as we train only small models. But R1 and R2 tend to form a conflict: the more natural a data source is, the less separable its domains tend to be. We considered different options, listed in Table~\ref{table:datamix-options}, to finally arrive at the routing testbed data mix. A discussion on the discarded alternatives is provided in Appendix \ref{app:datamix-options}.

\begin{table}[t]
\begin{center}
\begin{small}
\begin{tabular}{llp{2.5cm}ccc}
\toprule
\# & Data Source & Domains & R1 & R2 & R3 \\
\midrule
1 & Mixture of Gaussians & Gaussians & $\surd \surd$ & $\times \times$ & $\surd$ \\
2 & Samples from context-free grammars & Distinct vocabs or grammars & $\surd \surd$ & $\times$ & $\surd$ \\
3 & Any human language text corpus & Languages & $\surd$ & $\surd$ & $\surd$ \\
4 & Any human language text corpus & Text genres & $\surd$ & $\surd$ & $\surd$ \\
5 & Text classification datasets & Classes & $\surd$ & $\surd$ & $\times$ \\
6 & Typical LLM training mix & \textit{unknown} & $\times$ & $\surd$ & $\surd$ \\
\bottomrule
\end{tabular}
\end{small}
\end{center}
\caption{Potential data sources for the routing testbed mix. See main text for discussion.}
\label{table:datamix-options}
\end{table}

\subsubsection{Data Mix}

To build the proposed routing testbed, we construct a data mix based on option 3 in Table~\ref{table:datamix-options}. 
The mix uses Wikipedia articles\footnote{\url{https://www.wikipedia.org/}, licensed under CC-BY-SA 4.0} as the source. As Wikipedia exists in many different languages, we can rely on languages to obtain distinct domains. To maximize domain separability, we opted to use languages with different scripts. The data mix has 8 domains: German, Korean, Japanese, Chinese, Hebrew, Thai, Hindi and Arabic portions of Wikipedia data. Using different scripts provides better separability but is limited in domain coverage. In this mix, each domain comprises $1/D$-th of the mix, such that they are all of equal importance to the model and balancing objectives do not conflict with expert specialization.

\subsubsection{Token Split}
\label{sec:split}

While our domain-separated data mix provides a strong prior for expert specialization, not all tokens are equally domain-specific. Generic tokens like digits, punctuation or named entities appear across all languages and cannot be meaningfully attributed to a single domain. Naively treating all tokens as domain-specific would force the reference router to make arbitrary assignments for these generic tokens, likely creating a suboptimal reference. We empirically observe this in Section~\ref{sec:oracle}.

To address that challenge, we split tokens into a domain-specific ($\mathcal{T}_D$) and a generic ($\mathcal{T}_G$) subset. We train a feed-forward network as a token classifier on the data mix using domains as target labels, achieving 79\% test accuracy (model details in \ref{app:hyperparameters}). We then use classification confidence as a proxy for domain-specificity: tokens with high-confidence correct predictions are assigned to $\mathcal{T}_D$, while incorrectly classified tokens and low-confidence predictions are assigned to $\mathcal{T}_G$. The reference router applies domain-based routing only to tokens in $\mathcal{T}_D$, while tokens in $\mathcal{T}_G$ are routed freely. We test various setups to determine the optimal confidence threshold for the split and discuss the results in Section~\ref{sec:oracle}.

\subsection{MoE Model}
\label{sec:model}

We pair our $D$-domain data mix with a transformer MoE model having $E$ experts and top-$k$ routing such that $D=E/k$. In our setup, we use 32 experts with top-4 routing on an 8-domain data mix. Since every token is routed to $k$ experts and domains have equal sizes, this matched setup allows distinct groups of $k$ experts to specialize in different domains. To enable fast experimentation at small scale and examine routing behavior in the simplest possible setup, our default model uses just a single MoE layer. This single-layer design acts as a bottleneck in the network where the routing testbed can provide more interpretable results. Training details as well as variations of this base model are discussed in Section~\ref{sec:experiments} with additional details in Appendix \ref{app:additonal_experiments}.

\subsection{Metrics}

Given the routing decisions provided by a learned router that assigns tokens to experts, we want to compute the ``accuracy" of these assignments against the reference router's decisions. For that purpose, we define a metric named \textbf{routing purity} as follows

\begin{equation} 
\text{Routing Purity} = \frac{1}{E} \sum\nolimits_{i}^E \frac{\max_{d \in D} f_i^d }{ f_i }
\end{equation}

where $f_i$ denotes the fraction of tokens routed to expert $i$ and $f_i^d$ the fraction of that subset that is also from domain $d$. It measures to what extent the tokens routed to an expert are all from the same domain, thereby quantifying expert specialization along domain boundaries.\footnote{Our purity metric differs from traditional accuracy by focusing on within-expert consistency rather than specific expert assignments. Tokens from the same domain should be routed together; which expert serves them is irrelevant.} Importantly, we compute the metrics only over domain-specific tokens $\mathcal{T}_D$. Routing purity is our key metric to quantify \textbf{expert specialization}.

Our second metric, \textbf{expert utilization}, captures the idea that parts of the MoE's total model capacity is "lost" if experts are not evenly utilized (i.e., when an expert only receives few tokens then it can be undertrained). It is based on the per-expert token load as follows:

\begin{equation} 
\text{Expert Utilization} = \sum\nolimits_{i}^E \min \left( f_i, \frac{1}{E} \right)
\end{equation}

Utilization of an expert is maximized if the routing is perfectly balanced as the over-utilization of an expert implies under-utilization of other experts. 

Both metrics have a range of $[1/E, 1]$. Since they rely on per-expert token loads ($f_i$), they vary significantly based on what scope of tokens is used for computation. We always compute them over the full global batch. Note also that routing purity relies on a token's assigned domain provided by the reference router and thus is specific to our routing testbed setup, whereas expert utilization can always be computed. As a third metric, we rely on \textbf{validation loss}, the cross-entropy loss on a held-out portion of the routing testbed data mix, to measure a model's quality as a language model.

\subsection{Reference Router Verification}
\label{sec:oracle}

\begin{figure}[t]
\begin{center}
\begin{tikzpicture}[scale=0.9]
\begin{axis}[
    seaborn, seaborn lines,
    xlabel={Token split ratio ($\frac{|\mathcal{T}_D|}{|\mathcal{T}_D \cup \mathcal{T}_G|}$)},
    ylabel={Validation Loss},
    ylabel style={yshift=-2pt},
    grid=major,
    xmin=0.0, xmax=1.0,
    ymin=2.868, ymax=2.895,
    width=0.60\textwidth,
    height=0.35\textwidth,
]

\addplot[
    only marks,
    color=gray, mark=*, mark size=2pt, opacity=0.3] table [x expr={0.5}, y=loss, col sep=tab] {
loss
2.8914641439914703
2.8769383430480957
2.894556015729904
2.8911951780319214
2.8768711388111115
2.8799716532230377
2.889299660921097
2.8737289905548096
2.8883080184459686
2.873633086681366
2.873160094022751
2.872511565685272
2.872937768697738
2.8741078078746796
2.8731316328048706
2.8881969451904297
2.8726850748062134
2.886481076478958
2.8833106458187103
2.873681277036667
2.8738733530044556
2.8740996718406677
2.885833293199539
2.874529242515564
2.875358283519745
2.87497341632843
2.882509857416153
2.883062422275543
2.87393718957901
2.873316705226898
2.8756117820739746
2.871990889310837
2.8734994530677795
2.8823329508304596
2.8785810470581055
2.875068098306656
2.8775289952754974
2.87315633893013
2.882146567106247
2.8750225603580475
2.8743490874767303
2.8729952573776245
2.874352812767029
2.874705523252487
2.87516051530838
2.874351054430008
2.87881401181221
2.872643083333969
2.878473222255707
2.877946674823761
2.8753804862499237
2.8748300075531006
2.8743761479854584
2.8793312311172485
2.8723713755607605
2.879322588443756
2.8795164823532104
2.8793822824954987
2.877570658922195
2.8747453093528748
2.884128898382187
2.8847339153289795
2.877158463001251
2.8863271474838257
2.880748897790909
2.8830465972423553
2.885700106620789
2.8726036846637726
2.881173610687256
2.882034480571747
2.8874523043632507
2.872152864933014
2.8844308257102966
2.8839304447174072
2.8821074664592743
2.886088490486145
2.883078783750534
2.8816482722759247
2.8821690380573273
2.872374951839447
2.8829703629016876
2.8896769881248474
2.884288638830185
2.8763135969638824
2.883550226688385
2.8853611648082733
2.8825842440128326
2.886585235595703
2.8874371349811554
2.881750136613846
2.879828244447708
2.8880715072155
2.879009246826172
2.876766264438629
2.8760613203048706
2.87621545791626
2.881348252296448
2.8770121932029724
2.8756191730499268
2.8731818199157715
};

\addplot[
    color=red3,
    mark=*,
    mark size=2pt,
    line width=1pt,
] table[x=in-scope, y=loss] {
in-scope loss
0.0000 2.8729
0.3205 2.8726
0.4359 2.8710
0.5000 2.8699
0.6000 2.8702
0.7000 2.8745
0.8000 2.8736
0.9000 2.8835
1.0000 2.8834
};

\end{axis}
\end{tikzpicture}
\hfill
\begin{tikzpicture}[scale=0.9]
\begin{axis}[
    seaborn, seaborn lines,
    ybar,
    bar width=0.6cm,
    xlabel={},
    ylabel={Validation Loss},
    ylabel style={yshift=-2pt},
    grid=major,
    width=0.48\textwidth,
    height=0.35\textwidth,
    symbolic x coords={All ref, Our, No ref},
    xtick={All ref, Our, No ref},
    xticklabels={Domain-specific Routing, Our Routing Testbed, Learned Routing},
    xticklabel style={font=\small, text width=1.6cm, align=center},
    enlarge x limits=0.4,
    ymin=2.868, ymax=2.895,
]

\addplot[fill=gray!30, draw=black, bar shift=0pt,
    error bars/.cd, y dir=both, y explicit]
    coordinates {(No ref, 2.8791) +- (0,0.0055)};

\addplot[fill=blue!30, draw=black, bar shift=0pt]
    coordinates {(All ref, 2.8834)};

\addplot[fill=red3!30, draw=black, bar shift=0pt]
    coordinates {(Our, 2.8699)};

\end{axis}
\end{tikzpicture}
\vskip -0.40in
\caption{(Left) Validation loss under reference routing at varying token splits into $\mathcal{T}_D$ and $\mathcal{T}_G$, with gray points showing the loss range for fully learned routing. (Right) Naive domain-specific routing (split ratio 1.0) performs worst, fully learned routing improves but is sensitive to tuning, while our reference routing (split ratio 0.5) achieves lowest loss.}
\label{fig:oracle}
\end{center}
\end{figure}
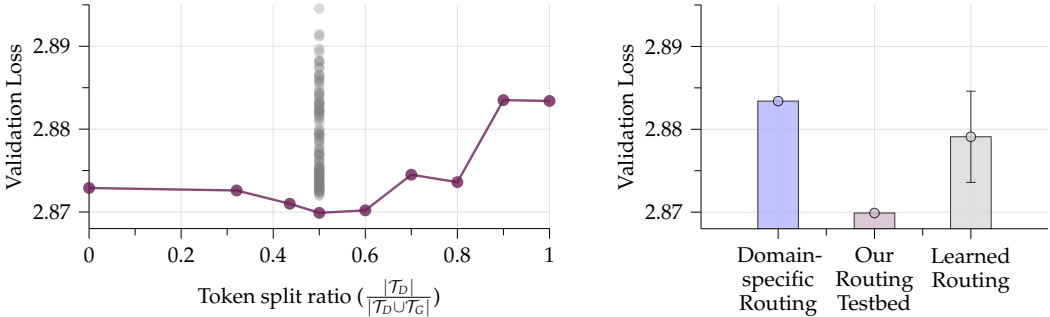

The reference router must provide an upper bound on performance to serve as a meaningful comparison target. We verify this by training MoE models where router logits are masked to enforce domain-based routing for domain-specific tokens:

\begin{equation} 
\mathbf{z}_{ti} = 
\begin{cases}
    \mathbf{z}_{ti} & if \,\, t \in \mathcal{T}_G \,\, \lor \,\, \lfloor i/k \rfloor = d \\
    -\infty & otherwise
\end{cases}
\end{equation}

It ensures domain-specific tokens of domain $d$ are restricted to $k$ experts specific to $d$.

Figure~\ref{fig:oracle} shows validation loss as we vary the split point between $\mathcal{T}_D$ and $\mathcal{T}_G$ based on classifier confidence thresholds (see Section~\ref{sec:split}). When the split ratio $\frac{|\mathcal{T}_D|}{|\mathcal{T}_D \cup \mathcal{T}_G|}=0$, no tokens are treated as domain-specific (all routing is learned); when the ratio is 1, all tokens follow domain-based routing. We find that forcing all tokens to follow domain-based routing yields suboptimal performance. The optimal split occurs at approximately 50:50, where roughly half the tokens (those with the highest classifier confidence) are treated as domain-specific. With this configuration, the reference router outperforms all learned routing methods tested, confirming it provides a valid upper bound for comparison.

\section{Load Balancing Experiments}
\label{sec:experiments}

In this section, we demonstrate the utility of our proposed testbed to identify better routing from small scale experiments. We study the impact of the token balancing dimensions introduced earlier (Section~\ref{sec:loadbalancing}) by systematically exploring the corresponding search space.

\subsection{Experimental Setup}

We explore a 3-dimensional token balancing search space: \textbf{1) Method:} LBL, EB, SH and BA. \textbf{2) Scope:} balancing over 1, 2, 8, 16 or all 64 sequences in the global batch by varying per-device micro batch size and aggregation across data-parallel ranks. \textbf{3) Strength:} for LBL and EB, we vary $\lambda \in {2^{-15}, 2^{-14}, \ldots, 2^0}$. In total, we evaluate 98 configurations.\footnote{Not all combinations are possible: SH and BA have no strength parameter; EB is applied only at global scope as originally proposed \citep{deepseekai2025deepseekv3technicalreport}; for LBL at global scope we use the approximation from \citet{qiu-etal-2025-demons}.}

Our default model is a 1-layer transformer MoE with 50M active parameters (270M total): 2048 hidden size, 16 attention heads and top-4 routing over 32 fine-grained experts with 1280 intermediate size. We use AdamW with a tuned learning rate of \num{2.44e-4} and train for 20k steps on approximately 10.5B tokens with global batch size 64 and sequence length 8192. We observe that our key metrics—routing purity and expert utilization—stabilize after 25-30\% of training, confirming this horizon is sufficient. See \ref{app:hyperparameters} for more details. For all results, we report utilization and specialization averaged over the final 100 training steps.

\begin{figure*}[t]
\begin{center}
\input{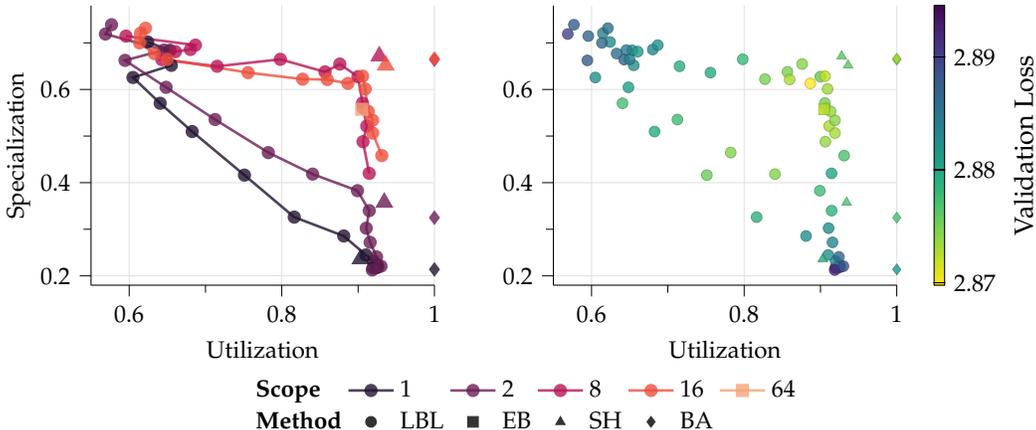}
\vskip -0.1in
\caption{Utilization-specialization trade-offs (left) and validation losses (right) captured by the routing testbed across MoEs with different balancing methods, scopes and strengths. Methods are indicated by marker style, scopes by color and strength decreases across points connected by a line (bottom-right to top-left, LBL only). As balancing is enforced less, utilization is traded off for specialization. Better trade-offs are possible when balancing over more token sequences. Validation loss correlates strongly with that trade-off.}
\label{fig:exp-base}
\end{center}
\end{figure*}

\subsection{Results}

\subsubsection{Balancing Scope is Crucial}

Figure~\ref{fig:exp-base} shows specialization, utilization and loss for the models trained across the balancing grid. At the smallest scope (1 sequence), LBL runs form an almost straight line from bottom-right to top-left as we decrease the strength, illustrating how less balancing trades off utilization for specialization. As the balancing scope increases, the conflict between these two goals is reduced and better trade-offs can be achieved. As shown on the right side of the figure, these trade-offs correlate strongly with validation loss, underlining the relevance of measuring that trade-off and the importance of balancing scope, a factor that has not received much attention in previous literature.

\subsubsection{Balancing Method is Minor Factor}

Figure~\ref{fig:exp-base} also compares the four balancing methods we tested. Since we cannot vary strength for methods other than LBL\footnote{We varied $\lambda_{EB}$ but found it to not control balancing strength smoothly. Instead, within a certain range of $\lambda_{EB}$, the degree of balancing remains constant. If above or below that range, EB quickly stops working completely and no longer achieves balance.}, they appear as individual points. As expected, BA achieves the best utilization as it always enforces perfect balancing, while SH and EB show trade-offs close to LBL at its best strength. Most notably, the differences between methods are minor compared to the effect of scope. At sufficiently large scope, all four methods work well. With that conclusion, we also confirm the finding of \citet{qiu-etal-2025-demons} showing LBL and EB to be on par when compared at the same scope.

\subsubsection{Token Dropping Prevents Specialization}
\label{sec:token-dropping}

\begin{figure*}[t]
\begin{center}
\input{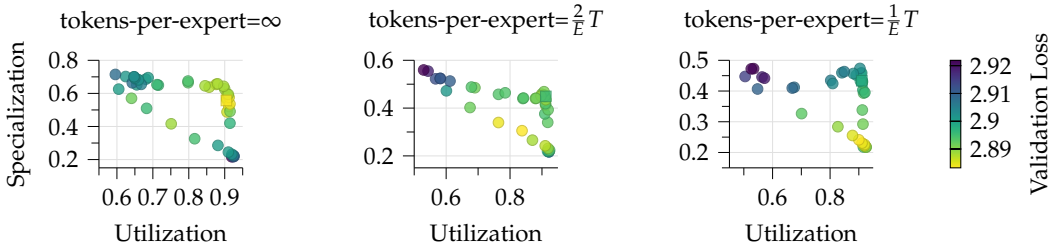}
\vskip -0.2in
\caption{Distribution of validation loss over the utilization-specialization landscape under different token-per-expert load constraints. Brighter is lower. The left plot matches Figure~\ref{fig:exp-base} but shows only a subset of the methods and scopes. As stricter per-expert load constraints are enforced (middle and right), best losses shift from top-right to bottom-right.}
\label{fig:exp-dropping}
\end{center}
\end{figure*}

A common approach in MoE training is to limit the local token imbalance to a certain capacity per expert and "drop" tokens above that limit by not processing them with any expert \citep{zoph2022stmoedesigningstabletransferable}. A capacity factor hyper-parameter typically defines this limit as a multiple of a perfectly balanced load of $\frac{1}{E}T$ tokens per expert. Our previous experiments applied no such limit, but Figure~\ref{fig:exp-dropping} shows results under this additional constraint.

We observe that such load constraints prevent specialization, as the highest obtained specialization values decrease with stricter constraints. In addition, the lowest validation loss shifts from coinciding with the best utilization-specialization trade-off (top-right) to strongest balancing at small scope (bottom-right). This is expected, as only strong local balancing can satisfy the (also local) token-per-expert load constraint and thereby avoid token dropping. The negative effect from token dropping is so significant that it more than offsets the benefits of specialization. This implies that the specialization benefits of larger scopes can only be realized if load constraints are loose enough to avoid substantial token dropping.

\subsubsection{Default Validation Loss is Insufficient}

We train the same 98 small-scale MoEs also on Dolma3 \citep{olmo2025olmo3}, a publicly available training data mix constructed for the OLMo 3 family of models. While the utilization-specialization trade-offs measured in our testbed correlate strongly with validation loss on our mix, the validation loss on Dolma3 does not. Thus, the default validation loss of a normal training mix cannot provide the same insights at small scale, underlining the value of the proposed routing testbed  (see Section~\ref{app:dolma_small_scale} for full experimental results).

\subsubsection{Findings Remain At Scale}
\label{sec:large-scale}

\begin{figure}[t]
    \centering
    \begin{tikzpicture}[scale=0.9]
\begin{axis}[
    seaborn, seaborn lines,
    xmode=log, 
    log basis x=2,
    xlabel={Scope (num. sequences)},
    ylabel={Validation Loss \\(rel. to scope=4)},
    ylabel style={yshift=-2pt},
    ylabel style={align=center},
    grid=major,
    xmin=3, xmax=96,
    ymin=0.988, ymax=1.0,
    width=0.7\textwidth,
    height=0.45\textwidth,
    legend style={font=\small,draw=none,fill=none},
    legend cell align=left,
    legend style={at={(1.35,0.9)}, anchor=north},
    cycle list name=tab10,
    every axis plot/.style={mark size=2pt, line width=1pt, mark=*},
    yticklabel style={/pgf/number format/.cd, fixed, fixed zerofill, precision=3},
]

\addlegendimage{empty legend}
\addlegendentry{\textbf{Validation Set}}

\addplot table[x=scope, y=loss] {
scope loss
4 1.
16 0.99752252
64 0.99668956
};
\addlegendentry{Overall}

\addplot table[x=scope, y=loss] {
scope loss
4 1.
16 0.9976832  
64 0.99709384
};
\addlegendentry{Common Crawl}

\addplot table[x=scope, y=loss] {
scope loss
4 1.
16 0.99757342 
64 0.99721571
};
\addlegendentry{olmOCR Science PDFs}

\addplot table[x=scope, y=loss] {
scope loss
4 1.
16 0.99564699 
64 0.98940314
};
\addlegendentry{Stack-Edu}

\addplot table[x=scope, y=loss] {
scope loss
4 1.
16 0.99492816 
64 0.99227991
};
\addlegendentry{FineMath 3+}

\addplot table[x=scope, y=loss] {
scope loss
4 1.
16 0.99397267 
64 0.98915826
};
\addlegendentry{arXiv}

\addplot table[x=scope, y=loss] {
scope loss
4 1.
16 0.9974063  
64 0.99552127
};
\addlegendentry{Wikipedia}

\end{axis}
\end{tikzpicture}
    \caption{Normalized validation losses after 2T tokens for MoE with 0.8B active (9.6B total) parameters. We can see clear improvements in performance with an increased balancing scope across all validation sets. Black shows validation loss across the full mix.}
    \label{fig:large_scale}
\end{figure}
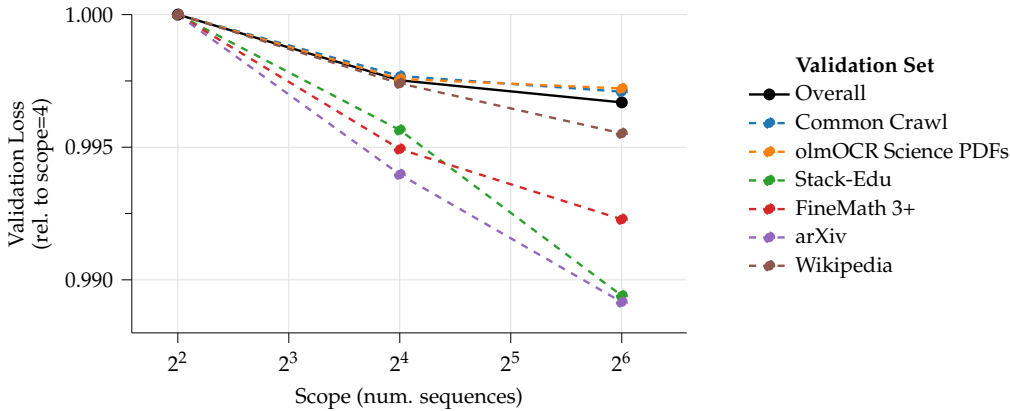

To verify the impact of balancing scope at scale, we train larger MoEs, also using Dolma3 as the data mix. The model configuration is as follows: 2048 hidden dimension, 18 layers, top-4 routing over 128 experts in every other layer with 1280 intermediate size, resulting in 0.8B active and 9.6B total parameters (see \ref{app:hyperparameters} for full details). This is 35x larger than the models trained in the routing testbed. We train all models using load balancing loss with auxiliary loss weight \num{2e-3}, global batch size 2240 and for 2T tokens. Figure~\ref{fig:large_scale} shows validation loss\footnote{We create custom held-out validation splits for the available mix subsets of Dolma3.} across balancing scopes of 4, 16, and 64 sequences. With this auxiliary loss weight, all runs achieve roughly 0.96 expert utilization. Despite identical utilization, larger scope yields substantially better validation loss, demonstrating that increased scope enables better utilization-specialization trade-offs, confirming our routing testbed findings at scale. Notably, the largest improvements occur on specialized validation sets such as Proof-pile Arxiv and Finemath compared to the Common Crawl subset, indicating that improved expert specialization particularly benefits evaluation on specialized domains. Absolute loss plots are shown in \ref{app:all_val_losses}.

\subsubsection{Less Ideal Expert-Domain-Pairings, Expert Granularity, Multiple Layers}

We show additional results for mismatched numbers of experts vs. domains, for varying expert granularity and for increasing the number of layers in appendices \ref{app:mismatch}, \ref{app:granularity} and \ref{app:depth}. Our general findings regarding the interaction of balancing scope, strength and method remain the same also under these additional conditions.

\section{Conclusion}

We introduced the MoE Routing Testbed, a framework for evaluating routing approaches in sparse MoE models at small scale while producing insights that generalize to larger models. The testbed pairs a carefully curated data mix with distinct, equally-sized domains with a reference router that provides an upper bound on performance. This design enables direct quantification of expert specialization through routing purity and expert utilization metrics, providing visibility into routing dynamics that standard perplexity metrics cannot capture.

Through extensive experiments, we identified balancing scope as the crucial factor in MoE training—more important than the choice of balancing method itself. When balancing is applied over sufficiently large token scopes, models can achieve favorable utilization-specialization trade-offs regardless of whether load balancing loss, expert bias or assignment algorithms are used. We further found that allowing sufficient tokens-per-expert at local scope is necessary for specialization to emerge, while strict token limits that trigger dropping negate these benefits. These findings remained consistent across variations in the number of domains and when scaling model depth or expert granularity. Standard validation losses offer no comparable insights at small scale. We further verified the role of scope generalizes to larger modes with a 0.8B active (9.6B total) parameter MoE (35x larger).

Our work demonstrates that small-scale experiments, when properly designed, can provide reliable insights into MoE routing behavior. The routing testbed methodology addresses a fundamental challenge in MoE research: the difficulty of distinguishing effective from ineffective routing configurations before committing substantial computational resources. By establishing clear ground truth through domain-separated data and reference routing, we enable principled comparison of routing approaches.

\section*{Acknowledgments}
We thank Saleh Soltan, Wael Hamza, Abhishek Kumar, Orchid Majumder, Thomas Gueudre and Shankar Ananthakrishnan for useful discussions and feedback on this work.

\bibliographystyle{colm2026_conference}
\bibliography{references}

\appendix
\section{Appendix}

\subsection{Balancing Methodologies}
\label{app:balancing_methods}

To demonstrate the utility of the developed routing test-bed, we compare the following token balancing approaches from the literature:

\textbf{Load Balancing Loss (LBL)}, used in GShard \citep{lepikhin2021gshard} and Switch Transformers \citep{fedus2022switch}, adds an auxiliary loss that penalizes uneven token distribution across experts. For a batch of $T$ tokens, let $f_i$ denote the fraction of tokens routed to expert $i$, and let $P_i = \frac{1}{T} \sum_{\mathbf{x}} g_i(\mathbf{x})$ denote the average routing probability for expert $i$. The balancing loss is defined as:

\begin{equation} 
L_{\text{LBL}} = E \cdot \sum\nolimits_i^E f_i \cdot P_i 
\end{equation}

This auxiliary loss is added to the primary loss with a weight $\lambda_{LBL}$: $L_{\text{total}} = L_{\text{CE}} + \lambda_{LBL} \cdot L_{LBL}$, enabling control over the balancing strength. LBL is the most commonly used method among recent MoE models.

\textbf{Balanced Assignment (BA)}, proposed in BASE layers \citep{lewis2021base}, formulates routing as a linear assignment problem: given a batch of $T$ tokens with routing scores $g$, find an assignment of exactly $T/E$ tokens to each expert that minimizes the sum $\sum g$. An auction algorithm is used to find a solution. This approach enforces a perfect balance across experts. 

\textbf{Sinkhorn Routing (SH)} was proposed as a more efficient alternative to BA \citep{clark2022}. It formulates expert assignment as an optimal transport problem. Given the routing logits $\mathbf{Z} \in \mathbb{R}^{T \times E}$, Sinkhorn iteratively normalizes the rows and columns of  $\mathbf{Z}$, converging to a state where each expert receives approximately $T/E$ tokens. The resulting matrix is used to select experts.

\textbf{Expert Bias (EB)} is the most recently proposed load-balancing method \citep{deepseekai2025deepseekv3technicalreport}. It adds an expert-specific bias to logits $\mathbf{z}' = \mathbf{z} + \mathbf{b}$ before the top-k selection step, enabling control over expert utilization by increasing/decreasing the bias. The biases are updated after each training step based on previous expert utilization with a update rate $\lambda_{EB}$, such that training converges to a balanced state over time. The routing scores used are without expert bias.

These four methods differ in several aspects: Only BA strictly enforces perfect load balancing, while the others tolerate some degree of imbalance. BA and SH affect the routing mechanism immediately, whereas LBL and EB have a delayed impact via parameter or bias updates. Among the four methods, only LBL relies on gradients for the balancing mechanism. With its auxiliary loss weight $\lambda_{LBL}$, the strength of the balancing can be controlled, which $\lambda_{EB}$ enables to some extent also for EB, while BA and SH do not have an explicit mechanism to adjust balancing strength.

\subsection{Balancing Scope and Sequence Packing}
\label{app:packing}

The $T$ tokens over which balancing methods are applied during MoE training are typically the product of a batch and a sequence dimension. The conflict between expert specialization and balancing at small batch-dimension scope (e.g. over a single sequence) only exists if there is no useful variation across the sequence dimension that allows effective specialization. If a training sequence is a single document, that tends to be the case. However, most documents are typically shorter than the full training sequence length and are packed together to increase training efficiency. Whether the group of documents that is packed together into a single sequence originates from a single data domain or multiple ones is therefore another important factor for balancing.

In this work, we exclusively use sequences of 8192 tokens packed from single domains, which leaves the batch dimension as the only one to study for balancing purposes. We emphasize that sequence-level domain mixing via packing would introduce only very limited additional flexibility: Documents of Dolma3 have 1650 tokens on average, such that 8192 token sequences contain on average 5 documents and could thus potentially enable specialization over 5 domains across the sequence dimension. The batch dimension, on the other hand, is orders of magnitude larger (we use global batch sizes of 64 and 2240) and therefore significantly more relevant for studying specialization under changing balancing scopes. In addition, the batch dimension provides a constant degree of diversity while the diversity within the sequence dimension would vary heavily from sequence to sequence.

\subsection{Data Mix: Explored Options}
\label{app:datamix-options}

We considered several data sources (Table~\ref{table:datamix-options}). Option 1 has been used in prior routing work \citep{dikkala-etal-2023-benefits} where input embeddings are sampled directly from Gaussian mixtures. While this provides controllable separated domains, the data is highly synthetic as it does not even consist of token sequences. Option 2 involves sampling from context-free grammars \citep{physics-of-llms} and moves closer to human language but remains highly synthetic. We tested this approach by reusing grammars from the referenced work but discarded it due to unusual training behavior with non-smooth loss curves. Option 5 uses text classification datasets which offer clearly labeled domains, though the dataset sizes and sample lengths are significantly smaller than required to enable strong model performance. In comparison, option 6 represents standard LLM training, where data is natural and vast but lacks clearly defined domains. Domains also likely vary significantly in size and importance, making their mapping to experts unclear.

\begin{table}[t]
\begin{center}
\begin{small}
\begin{tabularx}{\textwidth}{lXXX}
\toprule
& Small MoE (Base) & Larger MoE & Token Classifier \\
\midrule
Architecture & Llama-style Transformer Decoder & Llama-style Transformer Decoder & FFN (Transformer block w/o attention)\\
Output Layer & LM Head & LM Head & Classification Head \\
Active Parameters & 50M & 0.8B & 30M \\
Total Parameters & 270M & 9.6B & 30M \\
Num. Layers & 1 & 18 & 1 \\
Hidden Size & 2048 & 2048 & 2048 \\
Attention Heads & 16 & 16 & - \\
Query Groups & 16 & 8 & - \\
FFN hidden size & - & 5120 & 5120 \\
MoE Layers & 1 & 9 (alternating) & - \\
Num. Experts & 32 & 128 & - \\
Expert hidden size & 1280 & 1280 & - \\
Experts per Token & 4 & 4 & - \\
Sequence Length & 8192 & 8192 & 8192 \\
Global Batch Size & 64 & 2240 & 128 \\
Training Steps & 20k ($\approx$10.5B tokens) & 110k ($\approx$2T tokens) & 2 epochs \\
Optimizer & AdamW with $\beta=(0.9, 0.95)$, $\epsilon=$\num{1e-8} & AdamW with $\beta=(0.9, 0.95)$, $\epsilon=$\num{1e-8} & AdamW with $\beta=(0.9, 0.99)$, $\epsilon=$\num{1e-8} \\
Weight Decay & 0.1 & 0.1 & 0.01 \\
Max. Learning Rate & \num{2.44e-4} & \num{4e-4} & \num{2.4e-4} \\
Warmup Step & 2000 & 2000 & 73 \\
Decay Schedule & none & none & linear to 0 \\
\bottomrule
\end{tabularx}
\end{small}
\end{center}
\caption{Architecture and training hyperparameters for all models used in this work. For the small MoE we show the base setup, note some experiments vary selected parameters.}
\label{table:hyperparameters}
\end{table}

\subsection{Model Hyperparameters}
\label{app:hyperparameters}

Table~\ref{table:hyperparameters} shows the detailed model configuration and training hyperparameters used for the experiments reported in this work. All models are trained with PyTorch in bfloat16 precision on NVIDIA H100 GPUs. We train with a multi-lingual vocabulary of roughly 250k tokens. For the small MoE, we tuned the learning rate by sweeping $[-13.5,-13,-12.5,-12,-11,-10]$ in log-2 space for one point of the grid (LBL, 1 sequence, $\lambda_{LBL}=0.1$). The learning rate ramps up over the first 2000 steps and then stays constant. We observe that our metrics of interest, routing purity capturing expert specialization, and expert utilization, which we track for every step throughout the training, stabilize early, typically after about 25\%-30\% of the 20k steps horizon. We therefore conclude that a 20k step horizon is sufficient and also do not add a learning rate decay phase.

For the token classifier that we train to determine the token split into domain-specific and generic tokens, we use an architecture similar to the small base MoE, but make it a dense model. Additionally, we found that we also have to remove attention since the model will otherwise learn to label generic tokens with their correct domain as long as a domain-specific token is present in the attention window, resulting in near-perfect accuracy on the task (and no value for our use case of splitting the tokens). Hence, the resulting classifier is essentially just a feed-forward layer.

\subsection{Additional Experiments}
\label{app:additonal_experiments}

\subsubsection{Default Validation Loss Alone is Insufficient}
\label{app:dolma_small_scale}

Validation loss is typically used to evaluate model performance and compare training recipes or architectures. However, at small scales, validation loss can be unreliable due to noisy signals. This noise becomes more problematic when the data mix contains natural imbalances or when underlying factors interact in complex ways (see Section~\ref{sec:token-dropping}). We verify here the importance of our curated data mix for correctly assessing the relationship between specialization and utilization.

\begin{figure}[t]
    \centering
    \input{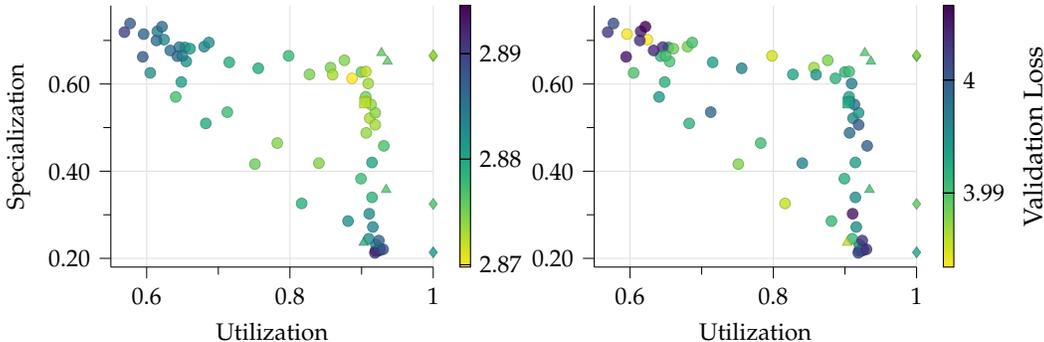}
    \vskip -0.2in
    \caption{We compare our small-scale models trained on the routing testbed data mix (left, matching Figure~\ref{fig:exp-base}) and Dolma3 (right). The color-coding shows validation loss on corresponding held-out validation sets (testbed vs. Dolma3) while the points are in both cases positioned based on testbed-measured specialization and utilization. While the trade-offs correlate strongly with validation loss on the testbed, this is not the case for Dolma3.}
    \label{fig:uninformative_val_loss}
\end{figure}

To illustrate this, we trained our small-scale models on Dolma3 \citep{olmo2025olmo3}, a realistic web-scale data mix with natural domain imbalances and compared this with the results from our testbed data mix. The results are shown in Figure~\ref{fig:uninformative_val_loss}. On this mix, the model achieving the best validation loss does not align with our specialization and utilization metrics. Across the 98 configurations of our grid, Spearman's rank correlation coefficient between a combined specialization-utilization-metric and validation loss is only -0.28 while it is at -0.80 on the testbed (Kendall's tau -0.18 vs. -0.58). It demonstrates why the MoE Routing Testbed requires a carefully balanced data mix with distinct, equally-sized domains. In such a setup, the best-performing model should naturally exhibit both high expert utilization and high expert specialization, making our metrics informative and interpretable.

\subsubsection{Dolma3 Validation Losses for Larger Model}
\label{app:all_val_losses}

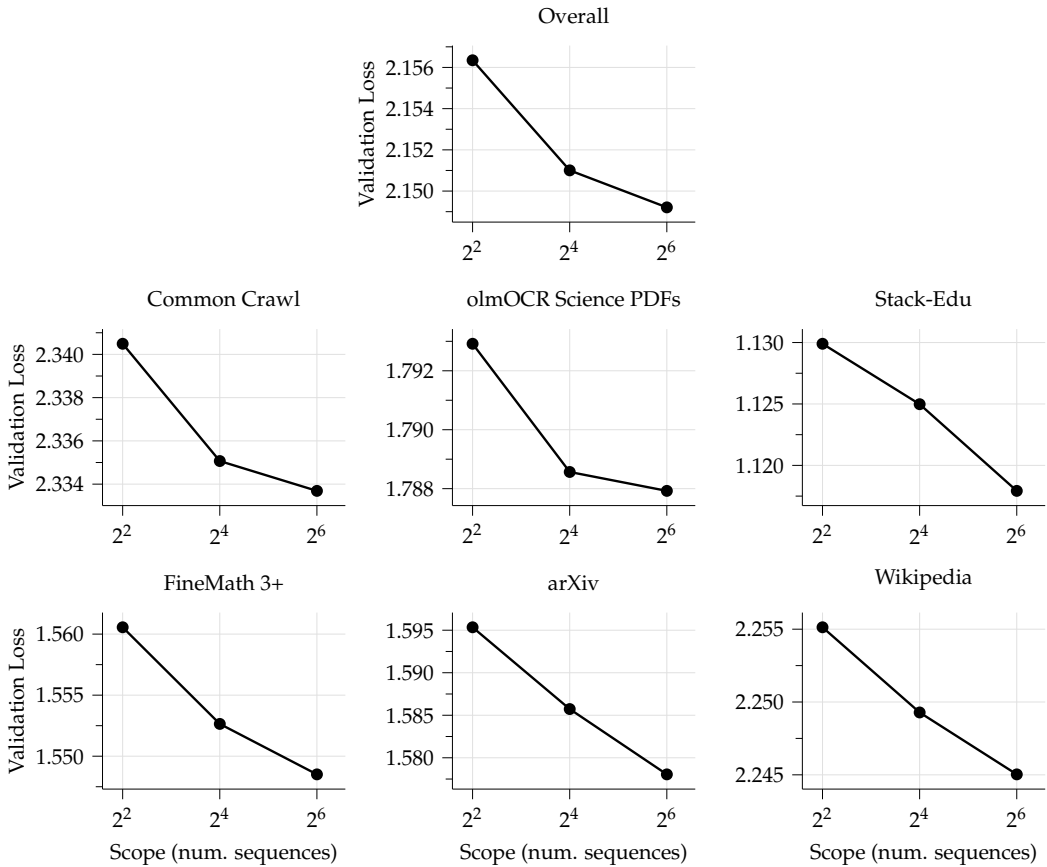
\begin{figure}[t]
    \centering
    \begin{tikzpicture}[scale=0.9]

\begin{axis}[
    title={Overall},
    seaborn, seaborn lines,
    xmode=log, 
    log basis x=2,
    ylabel={Validation Loss},
    ylabel style={yshift=-2pt},
    grid=major,
    xmin=3, xmax=96,
    enlarge y limits=0.1,
    width=0.37\textwidth,
    height=0.3\textwidth,
    xshift=0.37\textwidth,
    cycle list name=tab10,
    every axis plot/.style={mark size=2pt, line width=1pt, mark=*},
    yticklabel style={/pgf/number format/.cd, fixed, fixed zerofill, precision=3},
]
\addplot table[x=scope, y=loss] {
scope loss
4 2.1563481879432222
16 2.151005877472303
64 2.149209722851215
};
\end{axis}

\begin{axis}[
    title={Common Crawl},
    seaborn, seaborn lines,
    xmode=log, 
    log basis x=2,
    ylabel={Validation Loss},
    ylabel style={yshift=-2pt},
    grid=major,
    xmin=3, xmax=96,
    enlarge y limits=0.1,
    width=0.37\textwidth,
    height=0.3\textwidth,
    xshift=0.0\textwidth,
    yshift=-0.3\textwidth,
    cycle list name=tab10,
    every axis plot/.style={mark size=2pt, line width=1pt, mark=*},
    yticklabel style={/pgf/number format/.cd, fixed, fixed zerofill, precision=3},
]
\addplot table[x=scope, y=loss] {
scope loss
4 2.340490897977119
16 2.33506845290203
64 2.3336890554309004
};
\end{axis}

\begin{axis}[
    title={olmOCR Science PDFs},
    seaborn, seaborn lines,
    xmode=log, 
    log basis x=2,
    grid=major,
    xmin=3, xmax=96,
    enlarge y limits=0.1,
    width=0.37\textwidth,
    height=0.3\textwidth,
    xshift=0.37\textwidth,
    yshift=-0.3\textwidth,
    cycle list name=tab10,
    every axis plot/.style={mark size=2pt, line width=1pt, mark=*},
    yticklabel style={/pgf/number format/.cd, fixed, fixed zerofill, precision=3},
]
\addplot table[x=scope, y=loss] {
scope loss
4 1.7929166555404663
16 1.788565993309021
64 1.7879246473312378
};
\end{axis}

\begin{axis}[
    title={Stack-Edu},
    seaborn, seaborn lines,
    xmode=log, 
    log basis x=2,
    grid=major,
    xmin=3, xmax=96,
    enlarge y limits=0.1,
    width=0.37\textwidth,
    height=0.3\textwidth,
    xshift=0.74\textwidth,
    yshift=-0.3\textwidth,
    cycle list name=tab10,
    every axis plot/.style={mark size=2pt, line width=1pt, mark=*},
    yticklabel style={/pgf/number format/.cd, fixed, fixed zerofill, precision=3},
]
\addplot table[x=scope, y=loss] {
scope loss
4 1.1298986673355103
16 1.1249802112579346
64 1.1179252862930298
};
\end{axis}

\begin{axis}[
    title={FineMath 3+},
    seaborn, seaborn lines,
    xmode=log, 
    log basis x=2,
    xlabel={Scope (num. sequences)},
    ylabel={Validation Loss},
    ylabel style={yshift=-2pt},
    grid=major,
    xmin=3, xmax=96,
    enlarge y limits=0.1,
    width=0.37\textwidth,
    height=0.3\textwidth,
    xshift=0.0\textwidth,
    yshift=-0.6\textwidth,
    cycle list name=tab10,
    every axis plot/.style={mark size=2pt, line width=1pt, mark=*},
    yticklabel style={/pgf/number format/.cd, fixed, fixed zerofill, precision=3},
]
\addplot table[x=scope, y=loss] {
scope loss
4 1.560558795928955
16 1.552643895149231
64 1.5485111474990845
};
\end{axis}

\begin{axis}[
    title={arXiv},
    seaborn, seaborn lines,
    xmode=log, 
    log basis x=2,
    xlabel={Scope (num. sequences)},
    grid=major,
    xmin=3, xmax=96,
    enlarge y limits=0.1,
    width=0.37\textwidth,
    height=0.3\textwidth,
    xshift=0.37\textwidth,
    yshift=-0.6\textwidth,
    cycle list name=tab10,
    every axis plot/.style={mark size=2pt, line width=1pt, mark=*},
    yticklabel style={/pgf/number format/.cd, fixed, fixed zerofill, precision=3},
]
\addplot table[x=scope, y=loss] {
scope loss
4 1.5953443050384521
16 1.585728645324707
64 1.5780479907989502
};
\end{axis}

\begin{axis}[
    title={Wikipedia},
    seaborn, seaborn lines,
    xmode=log, 
    log basis x=2,
    xlabel={Scope (num. sequences)},
    grid=major,
    xmin=3, xmax=96,
    enlarge y limits=0.1,
    width=0.37\textwidth,
    height=0.3\textwidth,
    xshift=0.74\textwidth,
    yshift=-0.6\textwidth,
    cycle list name=tab10,
    every axis plot/.style={mark size=2pt, line width=1pt, mark=*},
    yticklabel style={/pgf/number format/.cd, fixed, fixed zerofill, precision=3},
]
\addplot table[x=scope, y=loss] {
scope loss
4 2.255131244659424
16 2.249282121658325
64 2.2450311183929443
};
\end{axis}

\end{tikzpicture}
    \vskip -0.2in
    \caption{Validation losses on Dolma3 validation sets after 2T tokens for 0.8B active MoE.}
    \label{fig:all_val_losses}
\end{figure}

In Figure \ref{fig:all_val_losses} we show the absolute validation losses of all Dolma3 validation sets across various balancing scopes (4, 16, and 64 sequences). Increasing the scope results in clear and meaningful improvements in performance overall across all validation sets.

\subsubsection{Findings Remain Under Less Idealistic Expert-Domain-Pairings}
\label{app:mismatch}

\begin{figure*}[t]
\begin{center}
\input{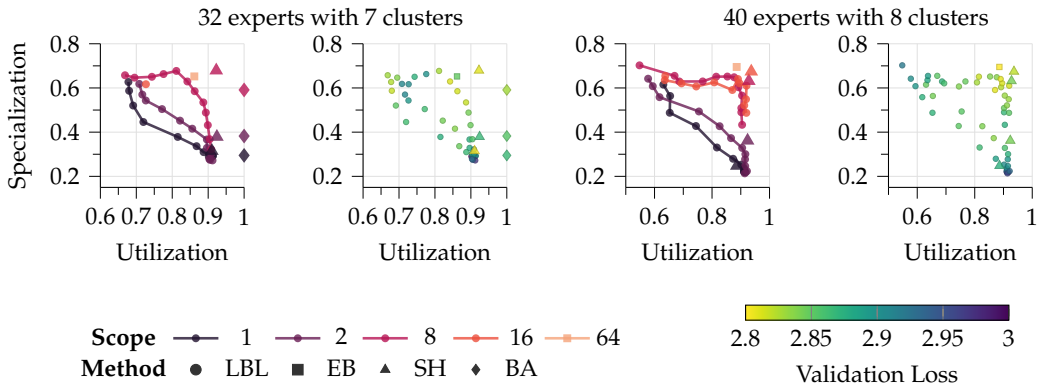}
\caption{Alternative testbed configurations with 32 experts trained on only 7 domains or 40 experts trained on 8 domains. 
Utilization-specialization and loss patterns remain consistent with the default 32 experts 8 domains setup.}
\label{fig:exp-mismatch}
\end{center}
\end{figure*}

The routing testbed is ideal as it uses exactly 8 equal-sized data domains to train 32 experts with top-4 routing, allowing distinct groups of 4 experts to specialize in each domain. To assess how much the results change as we deviate from this ideal state, we tested mismatched setups in which either the number of domains or experts is different. Figure~\ref{fig:exp-mismatch} shows similar plots as before for such settings. While the absolute values and exact shapes of the curves differ, the general observations still hold: balancing scope is the most important factor that allows specialization at high utilization whereas the difference between balancing methods is minor.

\subsubsection{Higher Expert Granularity Requires Less Token Balancing}
\label{app:granularity}

\begin{figure*}[t]
\begin{center}
\input{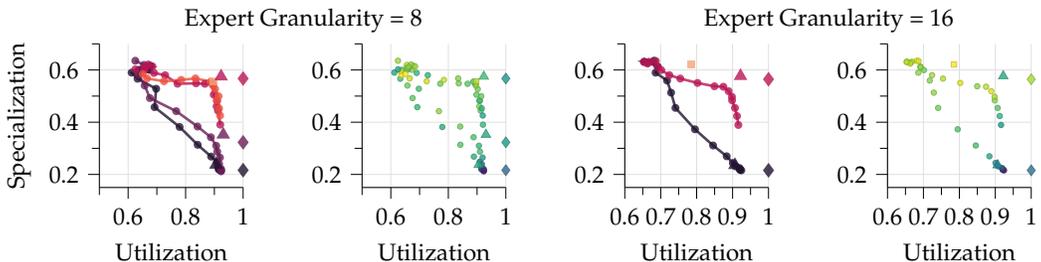}
\caption{As expert granularity increases, we observe a shift towards lower utilization for the best performing model. See Figure~\ref{fig:exp-base} for legend. Granularity 16 uses a smaller grid.}
\label{fig:exp-granularity}
\end{center}
\end{figure*}

Expert granularity, or fine-grained-ness, is a known dimension along which the performance of MoEs increases \citep{dai-etal-2024-deepseekmoe,krajewski2024scalinglawsfinegrainedmixture}. Increasing granularity refers to increasing the number of experts and top-k by a factor $g$ and dividing the intermediate size of each expert by $g$. The model can then choose from a larger pool of smaller experts. This modification keeps both total and active parameters constant but yields better performing models. Our base experiment in Figure~\ref{fig:exp-base} uses a granularity of 4. Figure~\ref{fig:exp-granularity} shows corresponding results for granularities 8 and 16. Again, the relationships between balancing methods, scopes and strengths remain consistent with previous results. Interestingly though, we notice that the utilization-specialization trade-off that results in the best validation loss shifts towards the top-left corner of the plot, which suggests that achieving high expert utilization becomes less important as the expert granularity increases.

\subsubsection{Specialization And Utilization Differ By Layer In Deeper Models}
\label{app:depth}

\begin{figure*}[t]
\begin{center}
\begin{tikzpicture}
    \begin{axis}[
    seaborn, seaborn lines,
    width=0.5\textwidth,
    height=0.5\textwidth,
    xlabel={Utilization},
    ylabel={Specialization},
    grid=major,
    xmin=0.35, xmax=1.0,
    ymin=0.15, ymax=0.8,
        legend style={font=\tiny,draw=none,fill=none},
    legend cell align=left,
    legend pos=north west,
     title={Scope = 1 Sequence},
]
\addlegendimage{empty legend}
\addlegendentry{\textbf{Layer}}
\addplot[
    color=blue, mark=*, mark size=1pt, line width=1pt,
    forget plot] table [x=utilization1,  y=specialization1, col sep=tab] {
utilization1	specialization1
0.925871753692627	0.2132948622107505
0.9284067392349244	0.2084783323109149
0.9219402074813844	0.2081430964171886
0.9236069679260254	0.2091209545731544
0.9167204856872558	0.2103785760700702
0.9203788757324218	0.21140598654747
0.9226092100143432	0.212500786781311
0.9216272592544557	0.2141222290694713
0.9231234073638916	0.2169524490833282
0.8849665641784668	0.2229413732886314
0.8576164960861206	0.2456843547523021
0.7543827772140503	0.2814249262213706
0.6610207796096802	0.3128997594118118
0.5321157455444336	0.3304855898022651
0.4833468437194824	0.3554587200284004
0.4497309207916259	0.3699285492300987
0.4248669147491455	0.3785974934697151
};

\addlegendimage{color=blue, no markers, line width=1pt}
\addlegendentry{2}

\addplot[
    color=cyan, mark=*, mark size=1pt, line width=1pt,
    forget plot] table [x=utilization3,  y=specialization3, col sep=tab] {
utilization3	specialization3
0.943870449066162	0.2059453442692756
0.926411485671997	0.2082086309790611
0.9246614933013916	0.2087638944387435
0.9329832077026368	0.2105011358857154
0.9280531883239748	0.2119891189038753
0.9271835803985596	0.2141722008585929
0.9225830554962158	0.2174534939229488
0.911452317237854	0.2270396135747432
0.8999065637588501	0.2410640805959701
0.8676461935043335	0.2791414111852645
0.8719253301620483	0.3220187023282051
0.8117434263229371	0.3594672203063965
0.8331541538238525	0.3767265722155571
0.799438214302063	0.4034570828080177
0.8209921598434449	0.4146862581372261
0.7762064933776855	0.4280588611960411
0.7939327001571655	0.4274094298481941
};

\addlegendimage{color=cyan, no markers, line width=1pt}
\addlegendentry{4}

\addplot[
    color=orange, mark=*, mark size=1pt, line width=1pt,
    forget plot] table [x=utilization5,  y=specialization5, col sep=tab] {
utilization5	specialization5
0.9403902292251588	0.2074152670800685
0.9323708772659302	0.2077999331057071
0.9336275577545166	0.210120067000389
0.940055513381958	0.2120384946465492
0.9331361293792724	0.2161772333085536
0.9295856714248656	0.2221570260822772
0.9183774948120116	0.2285530544817447
0.9169798851013184	0.2440983302891254
0.9106836795806884	0.271976138651371
0.8826997518539429	0.3411307260394096
0.8599218130111694	0.3840912118554115
0.8167513847351074	0.4178605228662491
0.842580509185791	0.4744998648762702
0.8319553852081298	0.5054956838488579
0.8267670154571534	0.5016395658254623
0.8149783849716187	0.538315835595131
0.8505292654037475	0.5295307785272598
};

\addlegendimage{color=orange, no markers, line width=1pt}
\addlegendentry{6}

\addplot[
    color=red5, mark=*, mark size=1pt, line width=1pt,
    forget plot] table [x=utilization7,  y=specialization7, col sep=tab] {
utilization7	specialization7
0.9341560125350952	0.2093815803527831
0.9352461576461792	0.2110042385756969
0.9295424222946168	0.2121224053204059
0.9374093055725098	0.2145622014999389
0.9259677410125732	0.218678468465805
0.9312247037887572	0.2267204523086547
0.9144040584564208	0.2371857866644859
0.9156793355941772	0.2617051899433135
0.9149420976638794	0.3043756529688834
0.8630174398422241	0.3731971830129623
0.8674981117248535	0.4945193350315094
0.8662020444869996	0.5812179952859878
0.8467660665512085	0.6267432659864426
0.8373715162277222	0.6476701200008392
0.8417585611343383	0.6527515649795532
0.8138997793197632	0.6573604315519332
0.8579957485198975	0.690796947479248
};

\addlegendimage{color=red5, no markers,  line width=1pt}
\addlegendentry{8}

\end{axis}
\begin{axis}[
    seaborn, seaborn lines,
    width=0.5\textwidth,
    height=0.5\textwidth,
    xshift=0.5\textwidth,
    xlabel={Utilization},
    grid=major,
    xmin=0.35, xmax=1.0,
    ymin=0.15, ymax=0.8,
        legend style={font=\tiny,draw=none,fill=none},
    legend cell align=left,
    legend pos=north west,
    title={Scope = 8 Sequence},
]

\addplot[
    color=blue, mark=*, mark size=1pt, line width=1pt,
    forget plot] table [x=utilization1,  y=specialization1, col sep=tab] {
utilization1	specialization1
0.9258798360824584	0.2137170657515525
0.922823452949524	0.2178641460835933
0.9213115215301514	0.2287997990846633
0.927857542037964	0.2506042763590812
0.9236276388168336	0.2827039569616317
0.9266167879104614	0.3105329349637031
0.9181723356246948	0.3287602424621582
0.919527578353882	0.3350651785731315
0.9213112115859984	0.3531617075204849
0.9045987606048584	0.3658846735954285
0.8489595413208008	0.3632156610488891
0.7808526992797852	0.360910901427269
0.6708784103393555	0.3402902826666832
0.5361173152923584	0.3375470533967018
0.4799750804901123	0.3584225162863731
0.4528375625610351	0.3986793875694275
0.4165072679519653	0.4160861909389496
};



\addplot[
    color=cyan, mark=*, mark size=1pt, line width=1pt,
    forget plot] table [x=utilization3,  y=specialization3, col sep=tab] {
utilization3	specialization3
0.9335355520248412	0.2308826081454753
0.9270159482955932	0.2566585905849933
0.931075119972229	0.3202789545059204
0.9211220026016236	0.3413548395037651
0.9179456949234008	0.3706592574715614
0.9073546648025512	0.3729723617434501
0.9098728895187378	0.3892861321568488
0.90138680934906	0.3988551720976829
0.8957255601882934	0.4005371287465095
0.8816872119903565	0.395975561439991
0.8572545766830444	0.4053324326872826
0.8366357088088989	0.4027756750583648
0.8020589351654053	0.4192671462893486
0.80903639793396	0.4206708908081055
0.8175537586212158	0.4273047313094139
0.792840552330017	0.4280221626162529
0.8046651840209961	0.4459932848811149
};



\addplot[
    color=orange, mark=*, mark size=1pt, line width=1pt,
    forget plot] table [x=utilization5,  y=specialization5, col sep=tab] {
utilization5	specialization5
0.9286727905273438	0.2457334585487842
0.9306557178497314	0.2749275684356689
0.9234292507171632	0.3451800122857094
0.92778000831604	0.4089938789606094
0.9156739950180054	0.4445170670747757
0.9207267761230468	0.4584468781948089
0.9024133443832396	0.4513584688305855
0.9027427196502684	0.4569707080721855
0.89644455909729	0.4643782764673233
0.8758345365524292	0.4836724579334259
0.8591685056686401	0.4792030155658722
0.8548465728759765	0.4897629410028458
0.8348371982574463	0.512971106171608
0.8222319602966308	0.5303698718547821
0.8137795925140381	0.524197393655777
0.8254563093185425	0.5505194395780564
0.8408133029937744	0.5271505981683731
};



\addplot[
    color=red5, mark=*, mark size=1pt, line width=1pt,
    forget plot] table [x=utilization7,  y=specialization7, col sep=tab] {
utilization7	specialization7
0.9290731906890868	0.45359568297863
0.9284916400909424	0.5005280256271363
0.9273695707321168	0.5698006868362426
0.924309539794922	0.6254593372344971
0.9221564531326294	0.6697521388530732
0.9221068382263184	0.6985720574855805
0.9102963209152222	0.7484842300415039
0.9117480754852296	0.7343477457761765
0.9071483373641968	0.7555297672748565
0.904779863357544	0.7235388100147248
0.862810468673706	0.6880224853754043
0.8648694038391114	0.7127424299716949
0.8550765752792359	0.7278621196746826
0.8178913593292236	0.6835334539413452
0.8152791261672974	0.6636668264865875
0.8341825246810913	0.6726187646389008
0.8429624080657959	0.678392294049263
};

\end{axis}

\end{tikzpicture}
\caption{Per-layer utilization-specialization trade-offs for 8-layer transformers with alternating dense and MoE layers. The obtained specialization and utilization values differ significantly between early and late layers. To illustrate the differences between layers, we separate scopes into different plots but instead combine all layers.}
\label{fig:exp-layers}
\end{center}
\end{figure*}
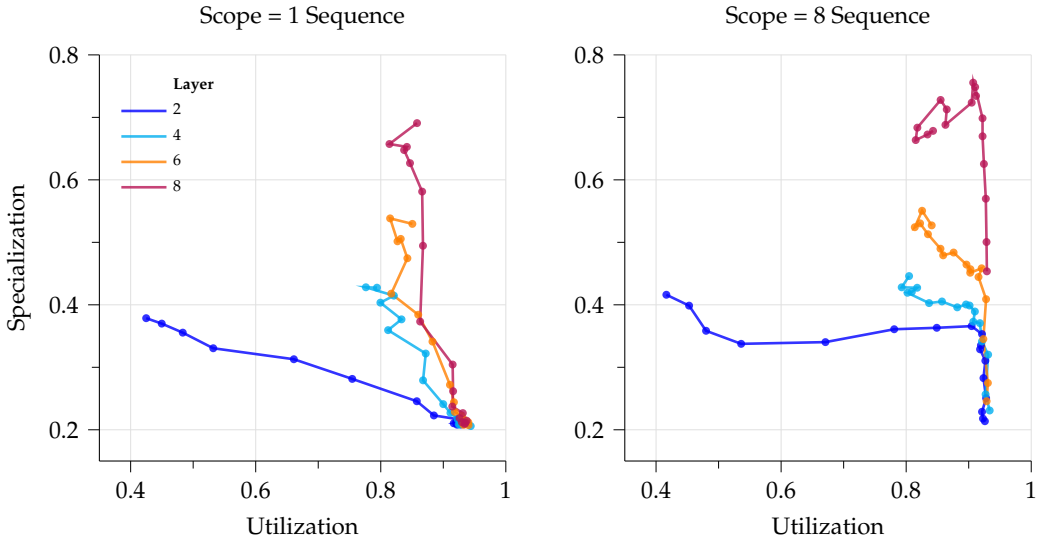

So far, all presented results were for our default 1-layer MoE model. We now increase the model depth to compare routing behavior across layers. Figure~\ref{fig:exp-layers} shows expert utilization and specialization for an 8-layer transformer with alternating dense and MoE layers (390M active, 1.27B total). Most notably, the value range for both utilization and specialization differs significantly across layers. In the first MoE layer, all models achieve only low specialization, whereas the last MoE layer has specialization higher than all previously shown results. At the same time, utilization varies widely across models in the first layer but is very high for every model in the final layer. This observation raises the question whether effective balancing should be layer-specific, e.g. by setting layer-wise balancing strengths or even using different scopes at different layers. We leave further exploration of this to future work.

\subsection{Extended Discussion: Throughput, Balancing and the impact on Specialization}

In this work we have emphasized the importance of specialization for training MoEs and, though we argue that MoEs can naturally specialize under relatively simple constraints (significant token scope), we have not clearly outlined the reasons as to why models have typically opted for more aggressive balancing strategies, overlooking a key detail. 

The emphasis on aggressive token balancing in MoE training stems from a key development: expert parallelism (EP). In traditional data parallelism (DP) with dense models, a global batch is sharded into local batches, each processed by a copy of the model on a separate rank. Expert parallelism modifies this for MoEs: rather than replicating all experts on every rank, each rank initializes only a subset of experts. Tokens are then communicated across ranks so that each expert processes tokens from the entire global batch, amortizing expert parameter costs across the data-parallel group. What remains is the communication cost of moving activations between ranks via all-to-all operations.

This architecture strongly favors balanced routing since if one expert receives most tokens while others sit idle, compute utilization collapses. Balancing has become central for two largely independent motivations:

\textbf{1. Training stability:} Without balancing, learned routers are prone to collapse: a small subset of experts attracts most tokens, starving other experts of gradient signal. These "dead experts" waste parameters and degrade model quality. Auxiliary balancing losses prevent this, ensuring all experts see sufficient signal for stable training.

\textbf{2. Throughput:} Expert parallelism is most efficient when each expert processes equal tokens. From a throughput perspective, perfect balance at every micro-batch is strictly optimal.

Both observations pushed the field toward defaulting towards aggressive balancing but mixing them has obscured an important fact: preventing dead experts requires only that experts receive meaningful gradient signal over training, while maximizing throughput demands near-perfect balance at every micro-batch. These are not the same requirement.

To further maximize throughput, many implementations introduced expert capacity limits: if an expert is oversubscribed, excess tokens are dropped. While this guarantees predictable compute, it's a severe overcorrection. Tokens are discarded not because they're uninformative, but because they violate a throughput constraint, actively preventing the specialization that motivates MoEs.

Both requirements conflict with specialization. Many balancing implementations enforce balance within each micro-batch. For high-quality data sources, a micro-batch may contain only a single sequence from one domain. Forcing tokens from a semantically coherent sequence to spread uniformly across all experts directly conflicts with specialization: that similar tokens should be routed together.

The combination of micro-batch balancing and capacity limits has inadvertently suppressed specialization. Compounding this, the field has lacked metrics to quantify specialization. In this work, we disentangle these effects through controlled experiments, measuring the contribution of each mechanism to performance and specialization. We show that balancing at larger scope preserves both training stability and throughput while enabling meaningful specialization, and that scope matters more than method choice.

\begin{figure}[htbp]
    \centering
    \includegraphics[width=1.0\textwidth]{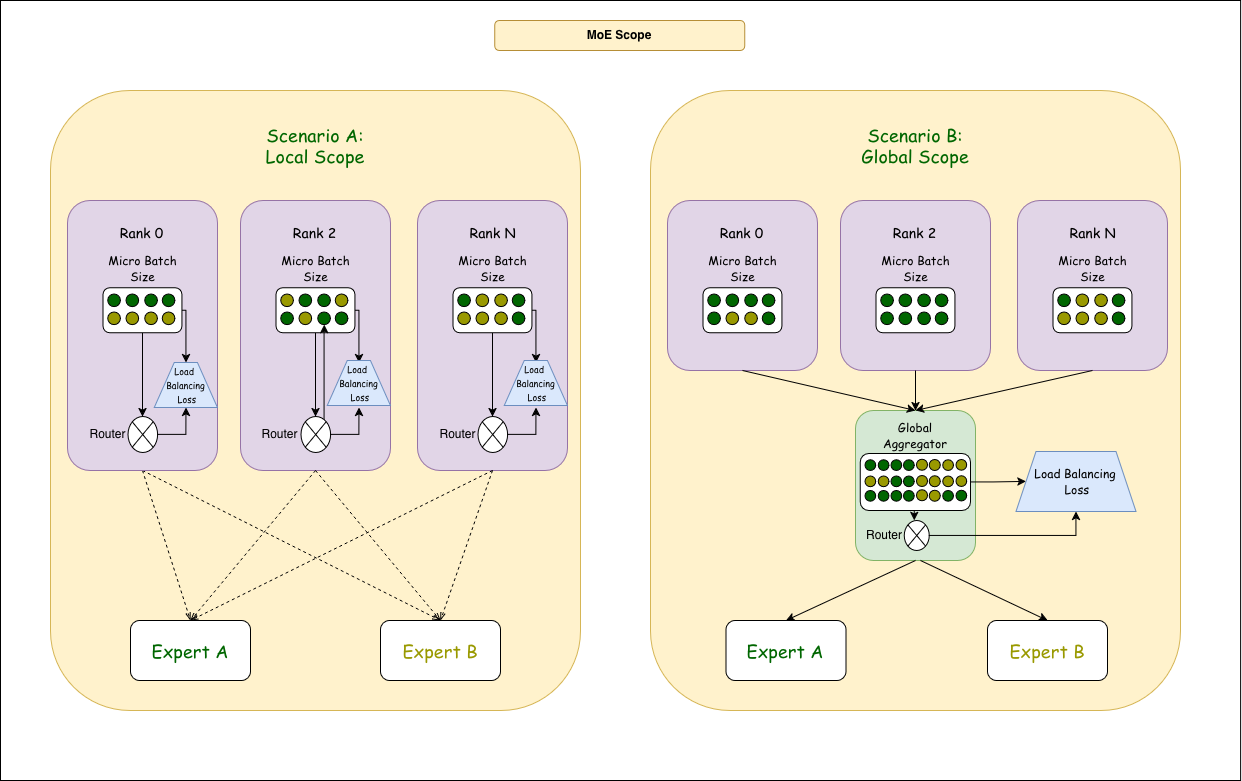}
    \caption{Load balancing loss aggregation over local scope versus global scope. Global, in this case, refers to computing loss across all DP ranks. }
    \label{fig:global_load_balancing_scope}
\end{figure}

\end{document}